\documentclass[final,5p,times,twocolumn,authoryear]{elsarticle}

\usepackage{amssymb}
\usepackage{amsmath}
\usepackage{graphicx}
\usepackage{amsfonts}
\usepackage[ruled,vlined,linesnumbered]{algorithm2e}
\usepackage{array}
\usepackage[caption=false,font=footnotesize,labelfont=rm,textfont=rm]{subfig}
\usepackage{textcomp}
\usepackage{stfloats}
\usepackage{xurl}
\usepackage{verbatim}
\usepackage{graphicx}
\usepackage{enumerate}
\usepackage{soul}
\usepackage{multicol}
\usepackage{orcidlink}
\usepackage{enumitem}
\graphicspath{{figures-pdf/}}
\usepackage{lipsum}
\usepackage{multirow,bigstrut}
\usepackage{tabularx}
\usepackage{threeparttable}
\usepackage{calc}

\usepackage{booktabs} 
\definecolor{lightgray}{gray}{0.9} 

\usepackage{color,xcolor}
\usepackage{colortbl}
\usepackage{diagbox}
\usepackage{ntheorem}
\usepackage{hyperref}
\usepackage{xcolor}
\usepackage{makecell}
\hypersetup{
 linktocpage=true,
 pdfborderstyle={/S/S/W 1},
 bookmarksopen=true,
 bookmarksnumbered=true,
}

\theoremseparator{:}

\newtheorem*{hyp*}{Hypothesis \protect\hypnumber} 

\newcommand{\hypnumber}{}

\newlength{\imagewidth}
\newlength{\threeimagewidth}
\newlength{\nineimagewidth}
\newlength{\vfigskip}
\newlength{\hfigskip}
\colorlet{myblue}{blue!90!gray}

    \newif\ifcomments
\commentstrue

\ifcomments
\newcommand{\BT}[1]{\textcolor{orange}{{#1}}}
\else
    \newcommand{\BT}[1]{}
\fi

\begin{document}

\begin{frontmatter}

\title{An Efficient Watermarking Method for Latent Diffusion Models via Low-Rank Adaptation and Dynamic Loss Weighting}

\author[1]{Dongdong Lin}
\author[1]{Yue Li}
\author[2]{Benedetta Tondi}
\author[3,4,5]{Kaiqing Lin}
\author[3,4,5]{Bin Li\corref{cor1}} 
\author[2]{Mauro Barni}

\address[1]{Xiamen Key Laboratory of Data Security and Blockchain Technology, Huaqiao University, Xiamen 361021, China}
\address[2]{Department of Information Engineering and Mathematics of the University of Siena, Italy}
\address[3]{Guangdong Provincial Key Laboratory of Intelligent Information Processing, Shenzhen University, Shenzhen 518060, China}
\address[4]{Shenzhen Key Laboratory of Media Security, Shenzhen University, Shenzhen 518060, China}
\address[5]{SZU-AFS Joint Innovation Center for AI Technology, Shenzhen University, Shenzhen 518060, China}
\cortext[cor1]{Corresponding author}  

\begin{abstract}
The rapid proliferation of Deep Neural Networks (DNNs) is driving a surge in model watermarking technologies, as the trained models themselves constitute valuable intellectual property. Existing watermarking approaches primarily focus on modifying model parameters or altering sampling behaviors. However, with the emergence of increasingly large models, improving the efficiency of watermark embedding becomes essential to manage increasing computational demands. Prioritizing efficiency not only optimizes resource utilization, making the watermarking process more applicable for large models, but also mitigates potential degradation of model performance. In this paper, we propose an efficient watermarking method for Latent Diffusion Models (LDMs) based on Low-Rank Adaptation (LoRA). The core idea is to introduce trainable low-rank parameters into the frozen LDM to embed watermark, thereby preserving the integrity of the original model weights. Furthermore, a dynamic loss weight scheduler is designed to adaptively balance the objectives of generative quality and watermark fidelity, enabling the model to achieve effective watermark embedding with minimal impact on quality of the generated images. 
Experimental results show that the proposed method ensures fast and accurate watermark embedding and a high quality of the generated images, at the same time maintaining a level of robustness aligned - in some cases superior - with state-of-the-art approaches. Moreover, the method generalizes well across different datasets and base LDMs. Codes are available at: \url{https://github.com/MrDongdongLin/EW-LoRA}.
\end{abstract}


\end{frontmatter}

\section{Introduction}
\label{intro}

In recent years, generative models have become increasingly popular for image synthesis, with Latent Diffusion Models (LDMs) emerging as one of the most prominent and effective approaches. Models like Stable Diffusion \cite{rombachHighResolutionImageSynthesis2022,podellSDXLImprovingLatent2023,esserScalingRectifiedFlow2024} and Diffusion Transformer \cite{peeblesScalableDiffusionModels2023a,crowsonScalableHighResolutionPixelSpace2024} have demonstrated outstanding performance, being capable of generating high-quality images with impressive realism, leading to commercial applications such as LiblibAI\footnote{LibLibAI: \url{https://www.liblib.art}} and SeaArt\footnote{SeaArt: \url{https://www.seaart.ai/zhCN}}. These models typically require extensive data and substantial computational resources to train, which has prompted the development of tools to prevent unauthorized uses and to safeguard intellectual property rights \cite{xueIntellectualPropertyProtection2022,zhangDeepModelIntellectual2022,fernandezWhatLiesAhead,liUniversalBlackMarksKeyImageFree2023c,liuWatermarkingDiffusionModel2023,pengProtectingIntellectualProperty2023,linCycleGANWatermarkingMethod2024a,sarcevicCanGenThis2024}.

A direct approach to achieving model identification and ownership verification is through \textit{model watermarking}, which involves embedding an identity message (the watermark) within the model parameters.
The watermark must satisfy the principles of unobtrusiveness and robustness \cite{barniDNNWatermarkingFour2021}. Unobtrusiveness implies that the watermark should not interfere with the model's primary task or significantly degrade its performance. Robustness means that the watermark should withstand potential modifications, such as post-processing, fine-tuning, or other model modifications.
In particular, when it comes to the protection of LDMs in open and untrusted environments, the embedded watermark should be retrievable from the output without the need to feed it with a specifically crafted input (box-free watermarking \cite{surveyDeepNeural2021}). 
Without loss of generality, in the following, we broadly categorize the existing LDM watermarking methods into three main categories.

\begin{figure}[!tb]
    \centering
    \begin{minipage}[t]{\columnwidth}
    \centering
    \includegraphics[width=\columnwidth]{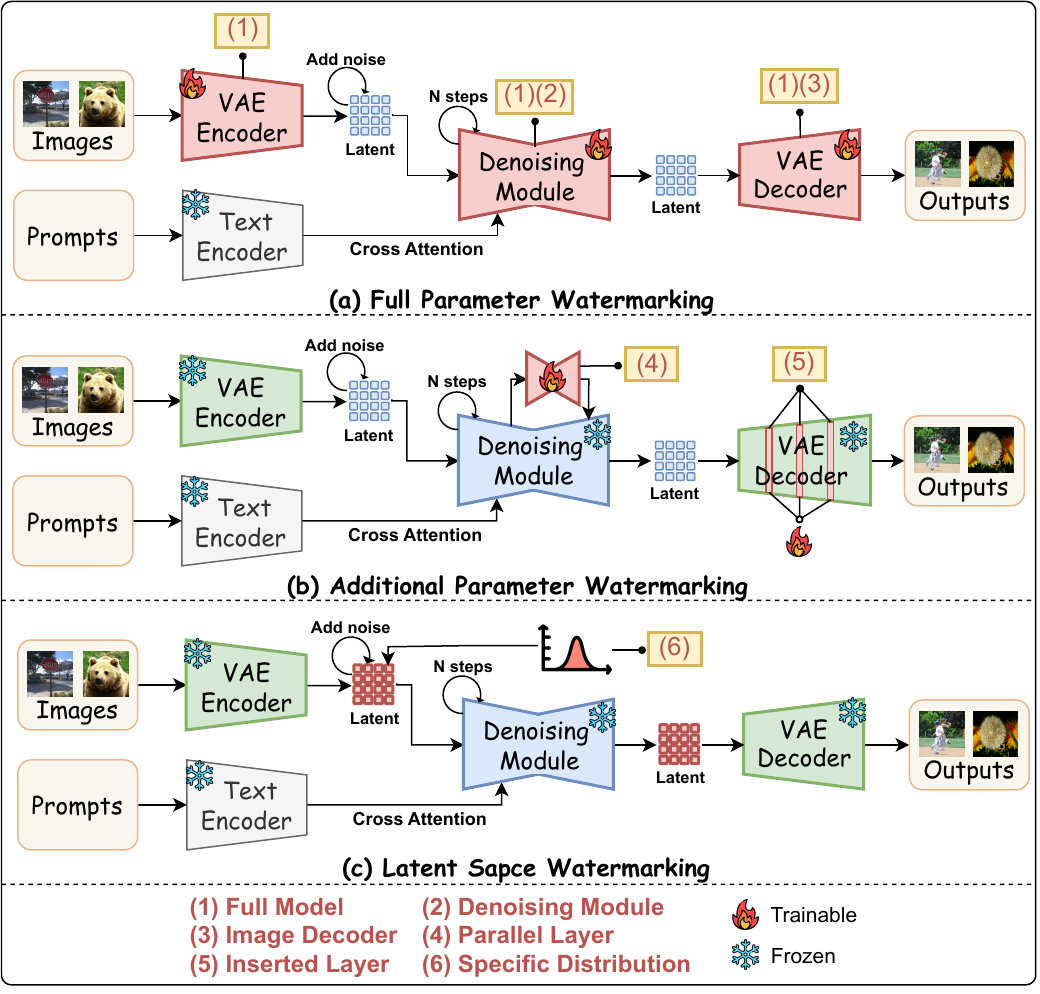}
    \end{minipage}
    \caption{A comparative overview of watermarking pipelines for latent diffusion models.} 
    \label{fig:pipelines}
\end{figure}

\emph{Full Parameter Watermarking (FPW)}: As shown in Fig.~\ref{fig:pipelines}-(1), an LDM can be watermarked by modifying its parameters through fully training or fine-tuning the model. A representative example is from \cite{zhaoRecipeWatermarkingDiffusion2023}, which embeds the watermark by jointly training Variational Autoencoder (VAE) and the denoising module (U-Net). 
However, as models increase in size, fully training an LDM becomes prohibitively costly due to the massive number of parameters involved.
For example, the large version of Stable Diffusion 3.5 \cite{esserScalingRectifiedFlow2024} contains 8.1 billion parameters, requiring days of training and substantial memory resources.
Some methods choose to embed the watermark by fine-tuning partial components of the LDM, such as the denoising module (see Fig.~\ref{fig:pipelines}-(2), as adopted by \cite{asnaniProMarkProactiveDiffusion2024, yuanWatermarkingStableDiffusion2024}), or VAE decoder (see Fig.~\ref{fig:pipelines}-(3), as adopted by \cite{fernandezStableSignatureRooting2023}). This watermarking pipeline reduces computational cost and memory usage while still enabling reliable ownership verification.

\emph{Additional Parameter Watermarking (APW)}: Another line of work embeds the watermark by introducing additional trainable parameters to the LDM. For example, methods proposed by \cite{rezaeiLaWaUsingLatent2025,ciWMAdapterAddingWaterMark2024} insert adaptive modules within the intermediate layers of the VAE decoder (see Fig.~\ref{fig:pipelines}-(4)). Alternatively, the method in \cite{fengAquaLoRAWhiteboxProtection2024} embeds the watermark by placing Low-Rank Adaptation (LoRA) \cite{huLoRALowRankAdaptation2021c} module in parallel with the existing layers of the LDM's denoising module (see Fig.~\ref{fig:pipelines}-(5)).
This design embeds the watermark without overwriting the original model weights, as the LoRA modules are applied in addition to the frozen parameters.

\emph{Latent Space Watermarking (LSW)}: Beyond the training-based methods, the method proposed by \cite{yangGaussianShadingProvable2024} explores training-free watermarking, where the watermark is embedded by sampling from specific noise distributions, thus avoiding model updates (see Fig.~\ref{fig:pipelines}-(6)). However, it suffers from watermark removal attacks \cite{mullerBlackBoxForgeryAttacks2025a, jainForgingRemovingLatentnoise2025, alamSaliencyAwareDiffusionReconstruction2025} and exhibits weaker robustness. By approximately inverting the diffusion process (e.g., DDIM \cite{songDenoisingDiffusionImplicit2020}), the adversary estimates the watermarked latent, identifies watermark-related patterns or regularities, and remove them prior to image decoding.

From Fig.~\ref{fig:pipelines}, we conclude that training-based methods consume substantial computational resources, whether they are retraining intrinsic modules in LDM or inserting plug-in layers with a large number of parameters (e.g., a linear layer in the U-Net). Such approaches are increasingly impractical given the current “bigbang” in parameter counts of generative models.

Existing LoRA-based watermarking methods, such as AquaLoRA~\cite{fengAquaLoRAWhiteboxProtection2024}, introduce modified LoRA modules into the denoising network with relatively high ranks for watermarking. However, these designs greatly increase the number of trainable parameters and computational cost. Moreover, applying LoRA to attention layers often leads to minor while visible semantic shifts in generated content, violating the unobtrusiveness requirement~\cite{barniDNNWatermarkingFour2021}.

In this work, we propose an \textbf{E}fficient \textbf{W}atermarking method for LDM based on \textbf{Lo}w-\textbf{R}ank \textbf{A}daptation (EW-LoRA), which efficiently and flexibly embeds watermarks with minimal impact on generation performance. Given a well-trained LDM, watermark embedding is achieved through parameter-efficient fine-tuning (PEFT), where additional parameters are introduced via LoRA. Furthermore, a dynamic loss weight scheduler (DLWS) is proposed to balance watermark fidelity and image quality. Unlike AquaLoRA, EW-LoRA applies LoRA to specific layers of the VAE decoder with a lower rank, introducing far fewer parameters and avoiding semantic distortion in latent space. The added parameters can be seamlessly merged into the original model without extra storage burden, ensuring both efficiency and unobtrusiveness.

In summary, the main contributions of this paper can be summarized as follows.

\begin{itemize}
\item We propose an efficient watermarking method for latent diffusion models based on low-rank adaptation, which significantly reduces the number of trainable parameters while maintaining high watermarking performance.
\item We propose a dynamic loss weight scheduler that accelerates the watermark embedding process and effectively reduces overall training time.
\item We conduct comprehensive evaluation metrics and validate the proposed method on multiple datasets and LDM architectures to verify the efficiency, robustness, and generalization of the proposed method.
\end{itemize}

The experiments we have run, show that our method significantly surpasses the state-of-the-art in terms of watermark accuracy and efficiency. We also show that EW-LoRA is robust against both image-level and model-level attacks. In particular, the robustness is similar - in some cases even superior - to the robustness of methods based on full parameter watermarking. Furthermore, EW-LoRA maintains high watermark fidelity across different datasets and LDM architectures.

\section{Related Works}
\label{sec:relatedworks}

\subsection{Model Watermarking}
\label{sec:modelWatermarking}

Since the pioneering work of \cite{uchidaEmbeddingWatermarksDeep2017}, who embedded watermarks into the weight space of Convolutional Neural Networks (CNNs), research on model watermarking has developed rapidly. Early methods mainly focused on CNNs in a white-box setting \cite{liSpreadTransformDitherModulation2021, darvishrouhaniDeepSignsendtoendwatermarking2019, zhangPassportawarenormalizationdeep2020}. Later, \cite{yuArtificialFingerprintingGenerative2021} proposed a black-box watermarking method for Generative Adversarial Networks (GANs), where watermarks are extracted from generated images via a frozen decoder—a strategy subsequently adopted in many works \cite{feiSupervisedGANWatermarking2022, linCycleGANWatermarkingMethod2024a}.

With the rise of diffusion-based generative models, latent diffusion models (LDMs) have become dominant in content generation, driving corresponding advances in watermarking techniques. As summarized in Section~\ref{intro}, the earliest attempts relied on \emph{Full Parameter Watermarking}, where the entire model was retrained on watermarked data \cite{zhaoRecipeWatermarkingDiffusion2023}, or partially fine-tuned, e.g., the VAE decoder \cite{fernandezStableSignatureRooting2023}, to reduce training overhead. 
Subsequent works explored more lightweight strategies, including \emph{Additional Parameter Watermarking} with adapter \cite{rezaeiLaWaUsingLatent2025} or LoRA modules \cite{fengAquaLoRAWhiteboxProtection2024}, and \emph{Latent Space Watermarking} without parameter updates \cite{yangGaussianShadingProvable2024}. 
While these directions demonstrated feasibility, they either incur high training costs with substantial trainable parameters or remain vulnerable to removal attacks, underscoring the need for more efficient and resilient watermarking strategies.

\subsection{Low-Rank Adaption Techniques}
\label{sec:lora}

Low-Rank Adaptation (LoRA) is a technique used for efficiently fine-tuning a large pre-trained model \cite{huLoRALowRankAdaptation2021c}. It works by adding small, trainable matrices to the model's parameter matrices, allowing for targeted updates without modifying the entire model. Given a parameter matrix $W \in \mathbb{R}^{m \times n}$, LoRA defines the adapted parameters as 
\begin{equation}\label{eq:lora}
    \tilde{W} = W + \Delta W, \quad \Delta W = \frac{\alpha}{r} \textcolor{orange}{BA},
\end{equation}
where $B,A$ are the trainable parameters, and
$B \in \mathbb{R}^{m \times r}$, $A \in \mathbb{R}^{r \times n}$, $r \ll \min(m,n)$.
and $\alpha$ is a scaling factor that controls the magnitude of the update. To ensure that the adaptation begins from the pre-trained model state, the parameters are initialized such that $\Delta W = 0$, typically by initializing $A$ with Gaussian noise and setting $B = 0$.
This formulation reduces the number of trainable parameters from $mn$ to $r(m + n)$, significantly lowering the cost of adaptation. In addition, the low-rank adaptation can be merged back into the original parameter matrix after training, removing no additional inference overhead.

There are many LoRA variants targeting different goals, mainly focusing on expressivity, optimization and efficiency. For example, \emph{LoHA} \cite{hyeon-wooFedParaLowRankHadamard2023} enhances expressivity by replacing a single low-rank update with a Hadamard product of two low-rank terms:
\begin{equation}\label{eq:loha}
    \tilde{W} = W + \Delta W, \quad \Delta W = \frac{\alpha}{r} (\textcolor{orange}{B_1 A_1} \odot \textcolor{orange}{B_2 A_2}).
\end{equation}

\emph{PiSSA} \cite{mengPiSSAPrincipalSingular2024} keeps LoRA’s form but initializes from the top-$r$ SVD of $W$ (often freezing the residual) to speed and stabilize training: write $W=U_r \Sigma_r V_r^\top + R$, set $B^{(0)} = U_r \Sigma_r^{1/2}$, $A^{(0)} = \Sigma_r^{1/2} V_r^\top$, then train parameter matrices $B,A$, i.e., 
\begin{equation}\label{eq:pissa}
    \tilde{W} = R+ \Delta W, \quad \Delta W= \frac{\alpha}{r} \textcolor{orange}{BA}.
\end{equation}

\emph{VeRA} \cite{kopiczkoVeRAVectorbasedRandom2024} targets parameter efficiency by sharing frozen low-rank bases $A,B$ across layers and training only per-layer scaling vectors $\mathbf b, \mathbf d$:
\begin{equation}\label{eq:vera}
    \tilde{W} = W+ \Delta W, \quad \Delta W = \operatorname{Diag}(\textcolor{orange}{\mathbf b}) B \ \operatorname{Diag}(\textcolor{orange}{\mathbf d}) A,
\end{equation}
where $\operatorname{Diag}(\textcolor{orange}{\mathbf b})$, $\operatorname{Diag}(\textcolor{orange}{\mathbf d})$ are the diagonal matrices with diagonals $\mathbf b$, $\mathbf d$. For a vector $\mathbf a=(a_1,\ldots,a_n)^\top$, we write
\[
\operatorname{Diag}(\mathbf a)=
\begin{bmatrix}
a_1 & 0   & \cdots & 0 \\
0   & a_2 & \cdots & 0 \\
\vdots & \vdots & \ddots & \vdots \\
0   & 0   & \cdots & a_n
\end{bmatrix}.
\]

For clarity we omit biases in the Eqs.~\eqref{eq:lora}, \eqref{eq:loha}, \eqref{eq:pissa} and \eqref{eq:vera}; if LoRA submodules include biases, they are absorbed into an effective $\Delta b$, yielding $y= (W+\Delta W)x + (b+\Delta b)$.

\section{Methodology}
\label{sec:methodology}

This section details EW-LoRA, covering the watermarking pipeline and its optimization strategy. We then formalize a threat model that encompasses both benign image post-processing operations and potential attacks the system may encounter.

\begin{figure*}[!tb]
    \centering
    \begin{minipage}[t]{0.9\textwidth}
    \centering
    \includegraphics[width=0.9\textwidth]{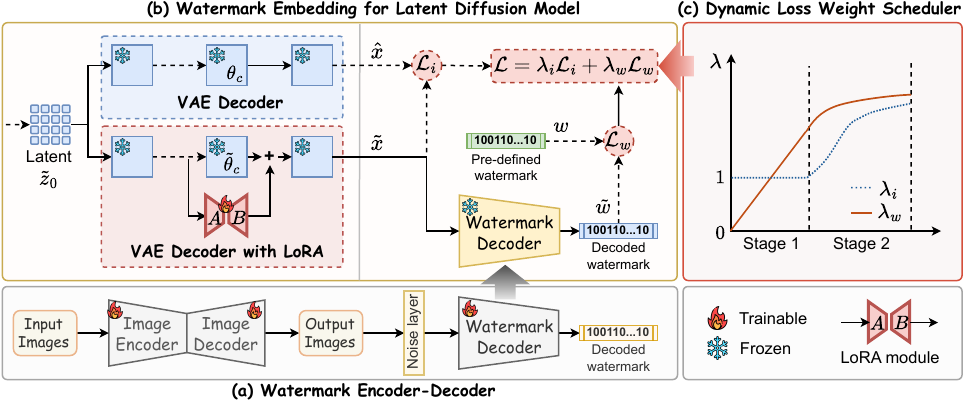}
    \end{minipage}
    \caption{The working flowchart of EW-LoRA. In a latent diffusion model, the parameters in the image decoder are selected for introducing LoRA. The LoRA parameters $A$ and $B$ are trained with a pre-trained watermark decoder. During training, a dynamic loss weight scheduler is proposed to balance the generative task and the watermark embedding task.}
    \label{fig:flowchart}
\end{figure*}

\subsection{Efficient Watermarking Method for LDMs}
\label{sec:method}

Consider the methods following the full parameter watermarking (FPW) and additional parameter watermarking paradigms (APW)\footnote{The paradigm of latent space watermarking (LSW) is excluded since it does not fit the parameter update approach and shows weak security under latent-based attacks \cite{mullerBlackBoxForgeryAttacks2025a, jainForgingRemovingLatentnoise2025, alamSaliencyAwareDiffusionReconstruction2025}}, model watermarking is achieved via parameter modifications. The parameter matrix $\tilde{\theta}$ of the watermarked model is obtained by mathematically adding a perturbation $\Delta \theta$ to the original model parameter matrix $\theta$, i.e.
\begin{equation}\label{eq.lora}
    \tilde{\theta} = \theta + \Delta \theta.
\end{equation}
Motivated by parameter efficiency in low-rank adaptation (i.e., LoRA), we parameterize the update in Eq.~\eqref{eq.lora} as $\Delta \theta = \frac{\alpha}{r}BA$, $B \in \mathbb{R}^{d \times r}$ and $A \in \mathbb{R}^{r \times d}$, $r \ll d$, which serves as a compact surrogate for the parameter increment, and $\alpha$ is a scaling factor that controls the magnitude of the parameter update.

As shown in Fig.~\ref{fig:flowchart}, the proposed method embeds the watermark by optimizing the additional parameters $B,A$ with a frozen watermark decoder $D_w$. The decoder extracts an $n$-bits code $\tilde w \in \{0,1\}^n$ from a watermarked image $I_w$, i.e., $D_w: I_w \mapsto \tilde{w}$. In prior works \cite{yuArtificialFingerprintingGenerative2021,linCycleGANWatermarkingMethod2024a,fernandezStableSignatureRooting2023,ciWMAdapterAddingWaterMark2024}, $D_w$ is typically pre-trained and then frozen, and the generative model is trained so that $D_w(I_w)$ matches the target bits; we follow this setup but update only $B,A$ with base parameters $\theta$ fixed.

\underline{\emph{Fewer Parameters}}: In parameter-efficient fine-tuning (PEFT) methods \cite{huLoRALowRankAdaptation2021c, hyeon-wooFedParaLowRankHadamard2023, mengPiSSAPrincipalSingular2024, kopiczkoVeRAVectorbasedRandom2024}, a set of lightweight trainable parameters is injected into the frozen backbone to adapt large-scale pre-trained models to downstream tasks. 
Considering that not all layers contribute equally to downstream adaptation or watermark embedding, we hypothesize that the number of additional parameters can be further reduced by selectively applying LoRA modules to only the most influential layers. Such selective embedding can preserve fine-tuning efficiency while maintaining or even improving the downstream fidelity and watermark capacity, especially when the modification of the feature flow in certain layers dominates the perceptual quality of the generated content.

Concretely, in an LDM we place LoRA only on selected layers of the decoder $D_I$ of the Variational Autoencoder (VAE). Given a denoised latent $\tilde{z}_0$, the output image is
\begin{equation}
        \tilde{x} = D_I(\tilde{z}_0; \tilde{\theta}), \hspace{0.5cm} \tilde{\theta}_c = \theta_c + \Delta \theta_c, \hspace{0.2cm} \tilde{\theta}_l = {\theta}_l  \text{ if $l \neq c$,}
\end{equation}
where
$\theta_c$ and $\tilde{\theta}_c$ are the parameters of the selected layers for introducing LoRA in the clean and the watermarked image decoder, and $\Delta \theta_c$ is replaced by the LoRA parameters. During training,
only the parameters in $\Delta \theta_c$ are trainable. In particular we let
\begin{equation} \label{eq:delta_theta_c}
    \Delta \theta_c = \frac{\alpha}{r} BA.
\end{equation}

It can be seen that Eq.~\eqref{eq:delta_theta_c} exactly meets the same form as in low-rank techniques (as in Eq.~\eqref{eq:lora}, \eqref{eq:loha}, \eqref{eq:pissa}, \eqref{eq:vera}). Denote $\hat{x}$ as the image generated from the clean image decoder, that is, $\hat{x} = D_I(\tilde{z}_0, \theta_c)$,
The total loss for training is
\begin{equation}
    \label{eq:loss}
    \mathcal{L} = \lambda_i\mathcal{L}_i (\tilde{x}, \hat{x})
    + \lambda_w\mathcal{L}_w (\tilde{w}, w),
\end{equation}
where
$\mathcal{L}_i$ is the generative loss and $\mathcal{L}_w$ is the watermark loss. For a watermarked image $\tilde x$, the decoded outputs $\tilde w = D_w(\tilde x)$, and $w$ is the target watermark. The weights $\lambda_i$ and $\lambda_w$ control the trade-off between generative fidelity and watermark detectability. We use Watson-VGG perceptual loss \cite{zhangUnreasonableEffectivenessDeep2018,czolbeLossFunctionGenerative2020}, also used in \cite{fernandezStableSignatureRooting2023}, for $\mathcal{L}_i$, and binary cross-entropy (BCE) loss for $\mathcal{L}_w$.

The watermark decoder $D_w$ is pre-trained with a differentiable noise layer to enhance robustness to common image post-processing attacks~\cite{feiSupervisedGANWatermarking2022,fernandezStableSignatureRooting2023}. During the subsequent LDM watermarking stage, $D_w$ is kept frozen and used to fine-tune the LoRA layers.

\begin{table}[h]
\centering
\caption{Loss configurations of representative LDM watermarking methods.}
\label{tab:loss_config}
\small
\setlength{\tabcolsep}{3pt}
\begin{tabular}{l l l}
\toprule
\textbf{Method} & \textbf{Loss weights} $(\mathcal{L}_i; \mathcal{L}_w)$ & \textbf{Ratio/Weights} \\
\midrule
Stable Sig.
& $\lambda_{\text{mse}}; \lambda_{\text{bce}}$ 
& $1.0;0.2$ \\

AquaLoRA
& $(\lambda_{\text{lpips}}, \mu_{\text{prvl}}); \lambda_{\text{bce}}$ 
& $(5, 0.5);1.0$ \\

WMAdapter
& $(\lambda_{\text{mae}}, \lambda_{\text{lpips}}, \lambda_{\text{vgg}}); \lambda_{\text{bce}}$ 
& $(0.2, 0.2, 0.08);1.0$ \\

LaWa
& $(\lambda_{\text{mse}}, \lambda_{\text{lpips}}, \lambda_{\text{adv}}); \lambda_{\text{bce}}$ 
& $(0.1, 1.0, 1.0);2.0$ \\
\bottomrule
\end{tabular}
\end{table}

\subsection{Dynamic Loss Weight Scheduler}

In most LDM watermarking methods, the training objective consists of a generative loss $\mathcal{L}_i$ and a watermark loss $\mathcal{L}_w$. These terms are typically combined with fixed weights, i.e., $\mathcal{L}=\lambda_i\mathcal{L}_i+\lambda_w\mathcal{L}_w$, and the weights are usually chosen empirically (refers to Table~\ref{tab:loss_config}). 
In practice, such fixed-weight choices can substantially affect convergence speed; \emph{empirically tuned values are often suboptimal and may lead to slower convergence}. When applied to large  generative models, this inefficiency becomes more pronounced, as the prolonged training time and repeated hyperparameter adjustments can significantly accumulate watermarking costs and computational overhead.
As shown in Section~\ref{sec:comparison}, our experiments confirm that fixed weights indeed slow convergence, especially when the values are selected from only a few pilot runs.

\underline{\emph{Faster Convergence}}: To accelerate the watermarking process, we introduce a dynamic loss weight tuning scheduler (DLWS). This scheduler balances the watermarking and generative tasks by dynamically adjusting the loss weights, $\lambda_i$ and $\lambda_w$, in each training iteration. The adjustments are based on the current batch's PSNR ($\rho_t$) and watermark extraction accuracy (BitACC, $\mu_t$). 
DLWS works by using a target PSNR ($\rho$) and a target BitACC ($\mu$), along with two counters, $c_i$ and $c_w$, which are both initialized to zero. This approach helps the watermark embedding process converge quickly by alternately adjusting the weights of the generative and watermark losses, as follows.
\begin{itemize}[itemsep=0pt]
    \item Initialize two success counters $c_i$ and $c_w$ to 0.
    \item If the current BitACC ($\mu_t$) is below the target ($\mu$) ($\mu_t<\mu$), we increase the watermark loss weight ($\lambda_w$) to prioritize improving the watermark's influence.
    \item If the current BitACC ($\mu_t$) meets or exceeds the target ($\mu$) ($\mu_t\geq \mu$), we increase the counter $c_w=c_w+1$. At this point, we check the PSNR:
    \begin{itemize}
        \item If the current PSNR ($\rho_t$) is below its target ($\rho$) ($\rho_t<\rho$), we increase the image loss weight ($\lambda_i$) to focus on improving the image quality.
        \item If the current PSNR ($\rho_t$) meets or exceeds its target ($\rho$) ($\rho_t\geq \rho$), we increment the counter $c_i=c_i+1$.
    \end{itemize}
    \item To continuously improve performance, if $c_w$ exceeds a stability target $p$, the target BitACC ($\mu$) is increased by a step size $s_\mu$. Similarly, if $c_i$ exceeds the taregt $p$, the taregt PSNR ($\rho$) is increased by a step size $s_\rho$. Then, reset the counters $c_i$ and $c_w$ to zero.
\end{itemize}

The staged operation of the DLWS algorithm effectively addresses the trade-off between the watermarking and generative tasks, achieving rapid convergence of the watermark embedding process by alternately adjusting the weights of the generative loss and the watermark loss. The algorithm operates in two stages, beginning by setting  $\lambda_i=1$ and $\lambda_w=0$ to prioritize optimal generative quality. In \textbf{Stage~1}, we focus on watermark robustness: as long as the watermark extraction accuracy (BitACC, $\mu_t$) is below the target value $\mu$, the algorithm continuously increases the watermark loss weight $\lambda_w$ to strengthen its influence. Once $\mu_t$ reaches the target, the system proceeds to \textbf{Stage~2}, where it performs dynamic weight balancing: loss weights are adjusted by increasing the weight corresponding to the metric (BitACC or PSNR) that falls below its predefined target (see Algorithm~\ref{algo:dlwt} for details). Figure \ref{fig:flowchart} visually demonstrates the variation of $\lambda_i$ and $\lambda_w$ across the training steps.

\begin{algorithm}[h]
    \caption{Dynamic Loss Weight Scheduler}
    \label{algo:dlwt}
    \KwIn{Current PSNR $\rho_t$ and BitACC $\mu_t$\; 
          \hspace*{3.1em}Current loss weights $\lambda_i$ and $\lambda_w$\; 
          \hspace*{3.1em}Target PSNR $\rho$; Target BitACC $\mu$\; 
          \hspace*{3.1em}Growth factor $\gamma$; Patience $p$\; 
          \hspace*{3.1em}Step sizes $s_\mu$, $s_\rho$}
    \KwOut{Updated $\lambda_i$, $\lambda_w$, $\mu$, $\rho$}
    $c_i \leftarrow 0$; $c_w \leftarrow 0$\;

    \If{$\mu_t < \mu$}{
        $\lambda_w \leftarrow \lambda_w + 2\gamma$ \tcp{Strengthen watermark}
    }
    \Else{
        $c_w \leftarrow c_w + 1$\;

        \If{$\rho_t < \rho$}{
            $\lambda_i \leftarrow \lambda_i + \gamma$\;
        }
        \Else{
            $c_i \leftarrow c_i + 1$\;
        }

        \If{$c_w > p$}{
            $\mu \leftarrow \min(\mu + s_\mu, 1.0)$\;
            $c_w \leftarrow 0$\;
        }

        \If{$c_i > p$}{
            $\rho \leftarrow \min(\rho + s_\rho, \inf)$\;
            $c_i \leftarrow 0$\;
        }
    }
\end{algorithm}

\section{Watermarking Requirements and Threat Model}\label{sec.problem}

To ensure that LDM watermarking methods are both functional and resilient, it is necessary to establish clear performance requirements and a realistic threat model. These definitions provide the foundation for designing, evaluating, and comparing watermarking schemes under practical deployment conditions.

\subsection{Watermarking Requirements}
\label{sec:metrics}

After the watermarking scheme is established, the watermark must satisfy a comprehensive and multidimensional evaluation to assess its overall performance. For the proposed efficient LDM watermarking method, a set of key performance requirements to ensure both functionality and practicality:

\begin{itemize}
    \item \emph{Efficiency}: The embedding process should achieve fast convergence with minimal computational overhead, supporting efficient model fine-tuning and scalable deployment.
    \item \emph{Accuracy}: The watermark must be extracted from generated results with high bit-wise correctness to ensure reliable identification and verification.
    \item \emph{Imperceptibility}: The watermark should not affect the perceptual quality or semantic fidelity of generated images. This ensures watermarked models can be deployed while maintaining the required generation quality.
    \item \emph{Robustness}: The watermark should remain detectable after common processing or attacks such as compression, filtering, or diffusion regeneration, which reflects its reliability under real conditions.
    \item \emph{Payload}: The watermark should carry sufficient information to support traceability and management.
\end{itemize}

\begin{table}[t]
\centering
\caption{Evaluation requirements and metrics for efficient LDM watermarking.}
\label{tab:requirements}
\small
\setlength{\tabcolsep}{3pt}
\begin{tabular}{m{2.0cm} m{1.8cm} m{4.5cm}}
\toprule
\textbf{Requirement} & \textbf{Metric} & \textbf{Description} \\
\midrule
\multirow{2}{*}{\emph{Efficiency}} 
  & Param & Number of trainable parameters involved in watermark embedding. \\ \cline{2-3}
  & AT-$x$ & Time to reach $x$\% extraction accuracy during watermark embedding. \\ \hline
\emph{Accuracy} & BitACC & Correctness of extracted watermark bits from generated outputs. \\ \hline
\emph{Imperceptibility} & PSNR, SSIM, LPIPS, SIFID & Visual similarity between clean and watermarked results, involves Peak Signal-to-Noise Ratio, Structural Similarity Index Measure, Learned Perceptual Image Patch Similarity, Single-Image Fréchet Inception Distance. \\ \hline
\emph{Robustness} & BitACC & Tests watermark detectability under content-level distortions such as compression, color enhancements, or regeneration. \\ \hline
\emph{Payload} & Bit length ($n$) & Information capacity for traceability or ownership proof. \\
\bottomrule
\end{tabular}
\end{table}

\subsection{Threat Model}\label{sec.threat_model}

In real-world applications, LDMs are exposed to a wide range of potential and evolving threats. It is essential that the model owner clearly specifies the deployment scenario, and carefully considers the diverse adversarial strategies that malicious users may adopt to overwrite or remove the watermark. An explicit and well-defined threat model enables the design of watermarking schemes that remain reliable and verifiable for ownership confirmation under realistic adversarial conditions.

In practice, LDMs are often distributed via platforms such as HuggingFace\footnote{Stable Diffusion Application: \url{https://huggingface.co/spaces/stabilityai/stable-diffusion}}, where users interact through prompts and only observe generated outputs. We assume that the generated images from the model may suffer from unauthorized use (box-free watermarking). 
Malicious users may apply post-processing attacks or may try to remove or overwrite the watermark to make the ownership verification fail or claim authorship \cite{fernandezStableSignatureRooting2023,ciWMAdapterAddingWaterMark2024,rezaeiLaWaUsingLatent2025}. 
The goal of the watermark embedder is to protect the intellectual property of the trained model.
Specifically, he aims to verify the ownership of the model by relying on the model's output in cases of unauthorized use following the release of the model.

\section{Experiments}
\label{sec:experiments}

In the experiments, we evaluate our proposed LDM watermarking method through a comprehensive metrics. Initially, we select the embedding layers for applying LoRA module; then, we test the EW-LoRA comparing to the state-of-the-art methods. Robustness performance in image post-processing and watermark overwriting and removal are tested. Finally, we show ablation on the chosen steps in DLWS algorithm.

\subsection{Selection of LoRA Embedding Layers}
\label{sec:ablation}

\begin{figure}[!t]
    \centering
    \begin{minipage}[c]{\columnwidth}
        \centering
        \includegraphics[width=\columnwidth]{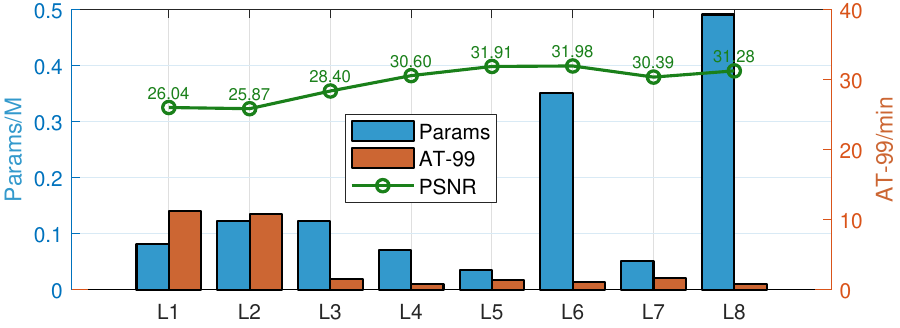}
    \end{minipage}
    \caption{Watermarking performance for applying LoRA in different network blocks in the VAE decoder of an LDM.}
    \label{fig:ablation_layers}
\end{figure}

\begin{figure}[t]
    \centering
    \begin{minipage}[t]{\columnwidth}
    \centering
    \includegraphics[width=\columnwidth]{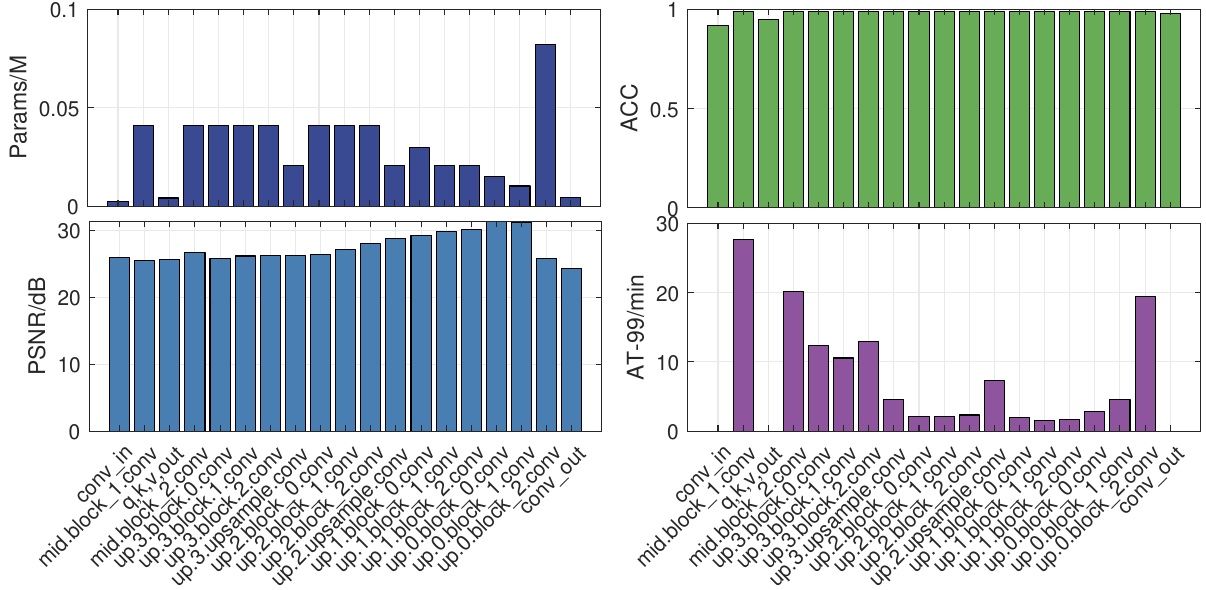}
    \end{minipage}
    \caption{Watermarking performance for applying LoRA in single layer in the VAE decoder of an LDM.}
    \label{fig:layer_metrics}
\end{figure}

To identify the most suitable layers for LoRA embedding, we conduct an ablation study on the VAE decoder of the LDM. The goal is to determine which components contribute most effectively to watermark embedding while maintaining image quality and parameter efficiency.
Specifically, two analysis settings are considered: a \emph{coarse-grained} setting that compares LoRA insertion across different decoder components, and a \emph{fine-grained} setting that examines single-layer insertion within the most promising region identified in the coarse analysis.

\emph{Experimental Setup}: In the coarse-grained setting, eight decoder components are examined as potential targets for LoRA insertion. As shown in Fig.~\ref{fig:ablation_layers}, we named these components as:
L1 – convolutional layers in the \texttt{mid.block};
L2–L5 – convolutional layers in the four upsampling blocks (\texttt{up.x.block}, $x=3,2,1,0$);
L6 – all convolutional layers across all \texttt{up.x.block}s;
L7 – the upsample convolutional layers; and
L8 – all convolutional layers throughout the VAE decoder.
The self-attention layers (\texttt{q}, \texttt{k}, \texttt{v}, and \texttt{out}) are excluded due to unstable convergence and poor watermark retention observed during training.
The LoRA hyperparameters are fixed as $r,\alpha=4,4$ in all configurations to ensure a fair comparison of parameter efficiency and watermark performance.

In the fine-grained setting, we further evaluate each individual layer within the \texttt{vae.decoder}, which contains 19 layers (as shown in Fig.~\ref{fig:layer_metrics}) in total. Each convolutional and attention-related layer is independently tested to measure its impact on watermark accuracy and image quality, allowing for a detailed assessment of layer sensitivity.

\emph{Results and Analysis}:
Figure~\ref{fig:ablation_layers} presents the coarse-grained results. Among all candidates, the configuration L5 — corresponding to the convolutional layers in \texttt{up.0.block} — achieves the best overall performance. It converges with fewer trainable parameters, faster AT-99 convergence, and higher image quality among all settings.
The fine-grained results, shown in Figure~\ref{fig:layer_metrics} show that layers such as \texttt{conv\_in}, \texttt{q}, \texttt{k}, \texttt{v}, \texttt{out}, and \texttt{conv\_out} fail to reach the 99\% watermark validation accuracy threshold. In contrast, the three convolutional layers within \texttt{up.0.block} achieve higher PSNR values with fewer trainable parameters, revealing a favorable trade-off between generative fidelity and watermark detectability.

Based on these analyses, \texttt{up.0.block.conv} is identified as the most effective and efficient target for LoRA-based watermark embedding. This layer offers the best balance among watermark accuracy, image quality and training efficiency and is therefore adopted as the default embedding position in all subsequent experiments.

\subsection{LoRA Parameter Configuration}

We further investigate the influence of the LoRA rank $r$ and scaling factor $\alpha$ (see Eq.~\ref{eq:delta_theta_c}) on the performance of the proposed method.
As summarized in Table~\ref{tab:ralpha}, increasing both $r$ and $\alpha$ generally enhances image quality and accelerates watermark convergence (lower AT-99), but at the cost of introducing more trainable parameters.
When $\alpha$ is increased while keeping $r$ fixed, AT-99 decreases and PSNR improves, with the configuration $(r,\alpha)=(4,32)$ outperforming $(4,4)$.
Conversely, increasing $r$ alone leads to a higher parameter count without significant gains in image quality, and only a marginal reduction in AT-99 (higher quality and lower AT-99 can be obtained by raising also  $\alpha$).
Therefore, to balance the requirements of efficiency, imperceptibility and robustness, we adopt $(r,\alpha)=(4,32)$ as the default configuration in the subsequent experiments.

\begin{table}[!t]
    \centering
    \caption{Watermarking performance when using different $r$ and $\alpha$.}
    \label{tab:ralpha}
    \small
    \setlength{\tabcolsep}{3pt}
    \begin{tabular}{llllll}
    \toprule
    $r,\alpha$ & \textbf{Params}$\downarrow$ & \textbf{BitACC}$\uparrow$ & \textbf{AT-99}$\downarrow$ & \textbf{PSNR}$\uparrow$  \\ 
    \midrule
    4,4 & 0.0353 & 0.991 & 2.518 & 31.733  \\ 
    8,8 & 0.0707 & 0.990 & 0.828 & 31.390  \\ 
    16,16 & 0.1413 & 0.990 & 0.794 & 32.446  \\ 
    32,32 & 0.2826 & 0.991 & 0.623 & 32.541  \\ \hline  
    4,8 & 0.0353 & 0.992 & 1.455 & 30.592  \\ 
    4,16 & 0.0353 & 0.991 & 0.803 & 31.027  \\ 
    4,32 & 0.0353 & 0.991 & 0.996 & 32.129  \\ \hline 
    8,4 & 0.0707 & 0.990 & 2.567 & 31.807  \\ 
    16,4 & 0.1413 & 0.992 & 1.546 & 32.342  \\ 
    32,4 & 0.2826 & 0.992 & 1.210 & 31.680  \\ 
    \bottomrule
    \end{tabular}
\end{table}

\subsection{Performance Evaluation and Comparison}
\label{sec:comparison}

\begin{table*}[!t]
	\renewcommand\arraystretch{0.92}
\caption{Performance comparison of LDM watermarking methods. 
The table compares state-of-the-art approaches, LoRA-based variants, and models with different watermark capacities. 
Metrics include number of the trainable parameters (Params), bit-wise accuracy (BitACC), achievement time (AT-99), visual quality (PSNR, SSIM, LPIPS, SIFID), 
and detection performance (TPR@0.1\%FPR). 
Subscripts with $\uparrow$ and $\downarrow$ indicate performance improvements after applying Dynamic Loss Weight Scheduler (DLWS) compared with the models trained with fixed weights. 
\textbf{Bold} and \underline{underlined} values denote the best and second-best results, respectively.}
\label{tab:comparison}
\small
\setlength{\tabcolsep}{5pt}
\centering
\begin{threeparttable}
\begin{tabular}{l|cc|cllll|>{\columncolor{cyan!10}}c>{\columncolor{cyan!10}}c}
\toprule
\textbf{Method (Bits)} & Params/M$\downarrow$ & BitACC$\uparrow$ & AT-99/min$\downarrow$ & PSNR$\uparrow$ & SSIM$\uparrow$ & LPIPS$\downarrow$ & SIFID$\downarrow$ & $\Delta_{\text{AT}-99}$$\downarrow$ & $\Delta_\text{PSNR}$$\uparrow$  \\ 
\midrule
\multicolumn{10}{l}{\textit{\textcolor{gray}{Comparison with State-of-the-art}}} \\
\midrule
Stable Signature (48) & 49.4900 & 0.989 & \underline{1.205}$_{\downarrow 24.880}$ & 29.281$_{\uparrow 0.280}$ & 0.815$_{\uparrow 0.009}$ & 0.061$_{\downarrow 0.012}$ & 0.116$_{\downarrow 0.015}$ & -24.880 & 0.280  \\ 
AquaLoRA (48) & 517.5586 & 0.959 & 682$^\dagger_{\downarrow 235}$ & 29.894$_{\uparrow 0.202}$ & 0.804$_{\uparrow 0.128}$ & 0.058$_{\downarrow 0.023}$ & 0.109$_{\downarrow 0.019}$ & -235 & 0.202  \\ 
WMAdapter (48) & 1.1978 & \textbf{0.999} & \textbf{0.973}$_{\downarrow 0.837}$ & 29.067$_{\uparrow 1.965}$ & 0.842$_{\uparrow 0.069}$ & 0.045$_{\downarrow 0.030}$ & 0.110$_{\downarrow 0.088}$ & -0.837 & 1.965  \\ 
LaWa (48) & 37.8943 & 0.970 & 3898$_{\downarrow 2339}$ & 32.808$_{\uparrow 0.548}$ & 0.861$_{\uparrow 0.019}$ & \textbf{0.011}$_{\downarrow 0.024}$ & \textbf{0.030}$_{\downarrow 0.055}$ & -2339 & 0.548  \\ 
\midrule
\multicolumn{10}{l}{\textit{\textcolor{gray}{LoRA Variants}}} \\
\midrule
Ours EW-LoRA (48) &  \cellcolor{cyan!10} \underline{0.0353} & \underline{0.997} & 2.931$_{\downarrow 0.816}$ & 32.543$_{\uparrow 0.206}$ & \underline{0.870}$_{\uparrow 0.001}$ & 0.025$_{\downarrow 0.004}$ & 0.093$_{\downarrow 0.031}$ & -0.816 & 0.206  \\  
Ours EW-LoHA (48) & \cellcolor{cyan!10} 0.0706 & 0.996 & 2.791$_{\downarrow 0.065}$ & \underline{34.073}$_{\uparrow 0.369}$ & \textbf{0.890}$_{\uparrow 0.009}$ & 0.018$_{\downarrow 0.001}$ & 0.082$_{\downarrow 0.006}$ & -0.065 & 0.369  \\ 
Ours EW-PiSSA (48) & \cellcolor{cyan!10} 0.0353 & 0.993 & 2.020$_{\downarrow 0.009}$ & \textbf{33.370}$_{\uparrow 1.797}$ & 0.860$_{\uparrow 0.025}$ & \underline{0.015}$_{\downarrow 0.005}$ & \underline{0.069}$_{\downarrow 0.011}$ & -0.009 & 1.797  \\ 
Ours EW-VeRA (48) & \cellcolor{cyan!10} \textbf{0.0015} & 0.994 & 38.148$_{\downarrow 55.905}$ & 31.516$_{\uparrow 2.001}$ & 0.825$_{\uparrow 0.048}$ & 0.025$_{\downarrow 0.019}$ & 0.104$_{\downarrow 0.120}$ & -55.905 & 2.001  \\ 
\midrule
\multicolumn{10}{l}{\textit{\textcolor{gray}{Capacity}}} \\
\midrule
Ours EW-LoRA (100) & \cellcolor{cyan!10} 0.0353 & 0.994 & 3.358$_{\downarrow 0.845}$ & 28.789$_{\uparrow 0.758}$ & 0.798$_{\uparrow 0.045}$ & 0.162$_{\downarrow 0.011}$ & 0.138$_{\downarrow 0.043}$ & -0.845 & 0.758  \\
Ours EW-LoRA (150) & \cellcolor{cyan!10} 0.0353 & 0.992 & 5.905$_{\downarrow 1.021}$ & 27.058$_{\uparrow 0.976}$ & 0.752$_{\uparrow 0.101}$ & 0.186$_{\downarrow 0.028}$ & 0.164$_{\downarrow 0.078}$ &-1.021 & 0.976  \\ 
\bottomrule
\end{tabular}
\begin{tablenotes}
    \footnotesize
    \item[$\dagger$] This represents AT-95 since AquaLoRA cannot reach 99\% validation accuracy.
\end{tablenotes}
\end{threeparttable}
\end{table*}

We compare the proposed EW-LoRA with several existing watermarking methods, including Stable Signature~\cite{fernandezStableSignatureRooting2023}, 
which belongs to the full parameter modification watermarking (FPW), and AquaLoRA~\cite{fengAquaLoRAWhiteboxProtection2024}, 
WMAdapter~\cite{ciWMAdapterAddingWaterMark2024}, and LaWa~\cite{rezaeiLaWaUsingLatent2025}, which fall into the additional parameter watermarking category (APW). 
To further evaluate the feasibility of the proposed method on different LoRA variants, we conduct experiments using LoHA~\cite{hyeon-wooFedParaLowRankHadamard2023}, PiSSA~\cite{mengPiSSAPrincipalSingular2024}, and VeRA~\cite{kopiczkoVeRAVectorbasedRandom2024}.
All models are trained on the COCO dataset~\cite{linMicrosoftCOCOCommon2014} using 6,000 randomly sampled images 
and evaluated on 1,000 randomly sampled images from the ArtELingo dataset~\cite{mohamedArtELingoMillionEmotion2022}. 
Each model is trained for one epoch with a batch size of four. 
When a model reaches the target validation bit-wise accuracy, we record the corresponding convergence time (AT-$x$) for comparison. Table~\ref{tab:comparison} gives the results of some LDM watermarking methods. Specifically, we compare the proposed EW-LoRA with representative state-of-the-art approaches and further evaluate its adaptability across various LoRA variants and watermark capacities.

\begin{figure}[!tb]
    \centering
    \begin{minipage}[t]{\columnwidth}
    \centering
    \includegraphics[width=\columnwidth]{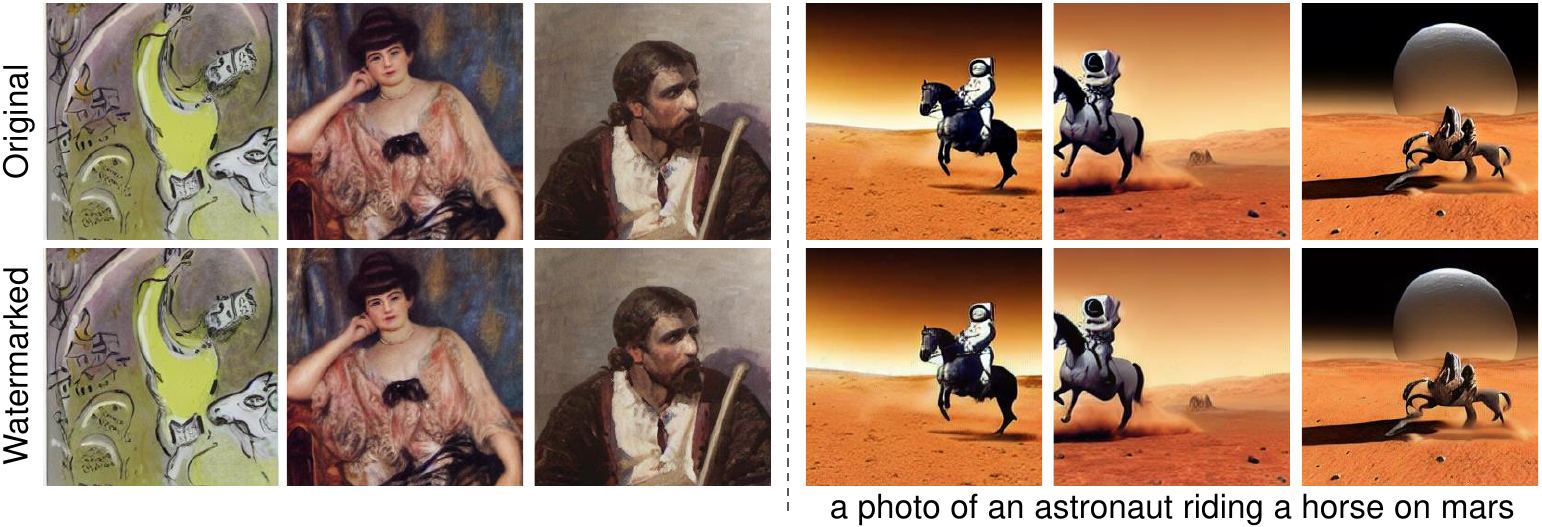}
    \end{minipage}
    \caption{Comparison between images generated by original SD v1.4 and our EW-LoRA watermarked SD v1.4: the leftmost six images and the rightmost six are generated, respectively, without and with a text prompt.}
    \label{fig:genimgs}
\end{figure}

\emph{Comparison with State-of-the-art}: 
We compare the proposed EW-LoRA with Stable Signature, AquaLoRA, WMAdapter, and LaWa. As shown in Table~\ref{tab:comparison}, EW-LoRA outperforms existing methods in both training efficiency and watermarking performance, achieving the fewest trainable parameters while maintaining high watermark accuracy and image fidelity.
The table also reports results with and without the proposed Dynamic Loss Weight Scheduler (DLWS), where each $\Delta$ value denotes the difference between the results with and without DLWS (i.e., $\Delta = \text{with} - \text{without}$). Negative $\Delta_{\text{AT-99}}$ values indicate that the model reach 99\% watermark validation accuracy faster after applying DLWS, while positive $\Delta_{\text{PSNR}}$ values demonstrate consistent improvements in image quality across the metric of PSNR.
In particular, EW-LoRA achieves high image quality with only 0.0353~MB of trainable parameters. Moreover, DLWS significantly reduces convergence time—for example, Stable Signature saves 24.88 minutes, AquaLoRA saves 235 minutes, WMAdapter saves 0.837 minutes, and LaWa saves 2339 minutes. The generative images in  Fig.~\ref{fig:genimgs} and Fig.~\ref{fig:genimgs_diff_models} further confirm that watermark embedding introduces negligible degradation in image quality.

\emph{LoRA Variants}: 
We evaluate the generality and effectiveness of the proposed EW-LoRA by integrating different LoRA variants, including LoHA, PiSSA, and VeRA, into the same embedding positions as EW-LoRA.
As shown in Table~\ref{tab:comparison}, EW-VeRA yields the fewest trainable parameters, while EW-PiSSA attains the highest image quality among all variants. Notably, all EW-LoRA variants surpass baselines such as Stable Signature, AquaLoRA, and WMAdapter in the quality of the generative images.
These results suggest that the proposed EW-LoRA design is compatible with diverse low-rank adaptation techniques and that reducing the number of trainable parameters helps mitigate degradation of generative image quality.

\emph{Capacity}:
For LDM watermarking methods such as Stable Signature, AquaLoRA, WMAdapter, and LaWa, the watermarking capacity primarily depends on the pre-trained watermark encoder–decoder architecture. In other words, their embedding capacity is inherently constrained by the design of the encoder–decoder network.
In our experiments, we evaluate EW-LoRA under different payload sizes, as summarized in Table~\ref{tab:comparison}. Even when the payload increases from 48 to 100 and 150 bits\footnote{Payloads of 100 and 150 bits use the StegaStamp~\cite{tancikStegaStampInvisiblehyperlinks2020} decoder as the pre-trained watermark decoder, since HiDDeN~\cite{zhuHiddenHidingdata2018} does not support such capacities.}, EW-LoRA maintains stable watermarking performance, exhibiting only minor decreases in PSNR and AT-99. This demonstrates its robustness to larger watermark capacities.

These results highlight two main advantages: (1) By exploiting the intrinsic low-rank adaptation property, EW-LoRA minimizes trainable parameters and memory usage compared with other LDM watermarking methods; and (2) The proposed DLWS effectively balances watermark robustness and generative fidelity, accelerating model convergence and improving overall training efficiency.

\subsection{Robustness}
\label{sec:robustness}

In this section, we evaluate the robustness of the watermarking methods against various attacks, including \emph{image post-processing attacks}, \emph{watermark overwriting attacks}, and \emph{watermark removal attacks}. Our standard EW-LoRA variant (Ours EW-LoRA (48) in Table~\ref{tab:comparison}) is compared with Stable Signature, AquaLoRA, WMAdapter, and LaWa.
For a fair comparison, all watermarking methods employ the same noise layer configuration during training. The \emph{noise layer} used in the watermark decoder consists of random cropping, resizing, and JPEG compression. Specifically, the crop or resize ratio is randomly set to either 0.3 or 0.7 with equal probability, followed by JPEG compression applied with a probability of 0.5, where the quality factor is randomly chosen as 50 or 80 with equal probability.

\emph{Image Post-processing Attacks}: 
Table~\ref{tab:postprocess} presents the robustness evaluation against common image post-processing operations, including center cropping (aspect ratios of 0.1 and 0.5), rotation ($25^\circ$), resizing (scaling factors of 0.3 and 0.7), brightness, contrast, and sharpness enhancement (all with adjustment factors of 1.5), and JPEG compression (quality factors of 80 and 50). These settings correspond to P1–P10 in the table. Together with the no-attack condition (P0), we have the results as shown in Table~\ref{tab:postprocess}. It can be seen that EW-LoRA consistently achieves top-tier robustness, ranking first or second across almost all attack types. This demonstrates that the proposed method effectively preserves watermark robustness under a wide range of post-processing distortions.

\emph{Watermark Overwriting Attacks}: 
For watermark overwriting attacks, we include two representative baseline attackers: StegaStamp~\cite{tancikStegaStampInvisiblehyperlinks2020}, a spatial-domain watermarking framework that embeds message bits into images through an encoder–decoder pipeline\footnote{StegaStamp: \url{https://github.com/ningyu1991/ArtificialGANFingerprints}. The number of embedded bits is set to $n=200$.}, and SSL-Watermarking~\cite{fernandezWatermarkingImagesSelfSupervised2022}, a zero-bit watermarking approach designed to detect ownership of images using self-supervised latent space\footnote{SSL-Watermarking: \url{https://github.com/facebookresearch/ssl_watermarking}.}.

To evaluate the resistance of the proposed method to watermark overwriting, we simulate attack scenarios where new watermarks are embedded into images generated by watermarked models using the two representative image watermarking schemes, StegaStamp and SSL-Watermarking.
Table~\ref{tab:attacks} reports the watermark BitACC extracted from both the overwritten images (i.e., images generated by the watermarked model and subsequently embedded with another watermark) and the original images generated by the same watermarked model.
As shown, EW-LoRA achieves BitACC values comparable to those of the best-performing WMAdapter (with only a 0.002 BitACC gap), demonstrating strong robustness against overwriting watermark and effective preservation of the original watermark under overwriting attacks.

\emph{Watermark Removal Attacks}: 
To evaluate the robustness of the watermarking methods against watermark removal attacks, we employ two learned image compression autoencoders, Bmshj2018~\cite{balleVariationalimagecompression2018} and Cheng2020~\cite{chengLearnedimagecompression2020}, which serve as representative encoder–decoder architectures capable of reconstructing images while potentially removing embedded watermarks.
We use the implementations provided in the \texttt{CompressAI.zoo} library\footnote{ConpressAI: \url{https://github.com/InterDigitalInc/CompressAI}}
 and vary the compression rates from ${1,2,3,4,5}$ for Bmshj2018 and from ${1,2,3,4,5,6}$ for Cheng2020.
 
\begin{figure}[!t]
    \centering
    \begin{minipage}[c]{\columnwidth}
        \centering
        \includegraphics[width=\columnwidth]{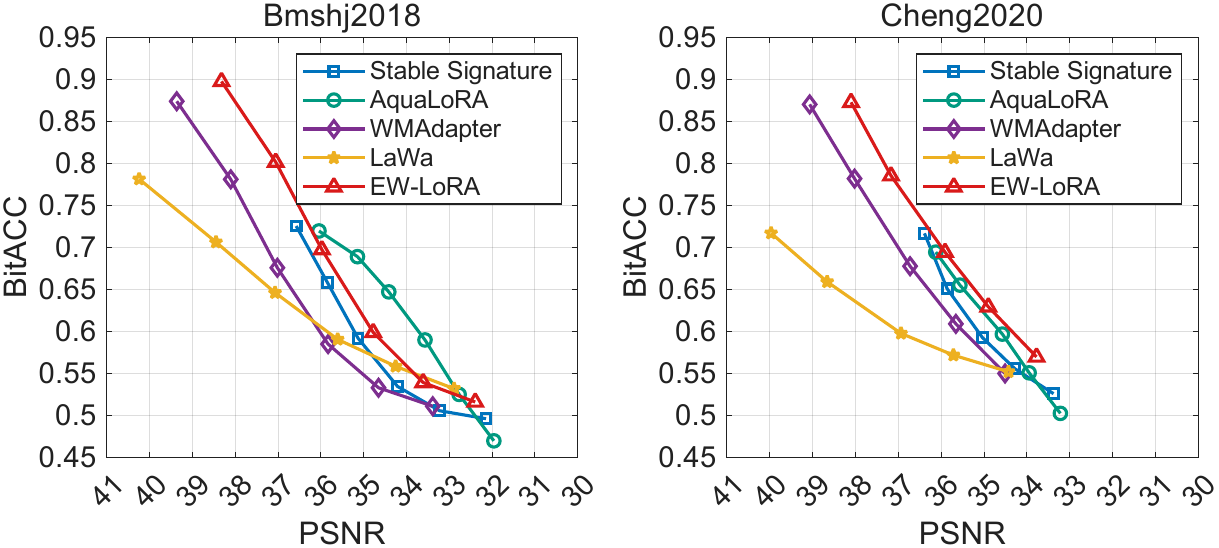}
    \end{minipage}
    \caption{Watermarking performance under watermark removal attacks.}
    \label{fig:watermark_removal}
\end{figure}

For watermark removal attacks, we assess robustness by passing the watermarked images through the above compression models.
The PSNR is computed between the original watermarked images and their reconstructed counterparts, while the BitACC is calculated on the reconstructed images to measure the watermark’s recoverability after compression.
As shown in Figure~\ref{fig:watermark_removal}, EW-LoRA demonstrates superior robustness compared to the other three methods, achieving consistently higher watermark BitACC under the same distortion levels (same PSNR) on the watermarked images (except for the case of AquaLoRA under the attack of Bmshj2018).
These results indicate that EW-LoRA effectively preserves watermark robustness even after severe compression-based reconstruction.

\begin{table}[t]
    \centering
    \caption{Watermarking performance comparison in the presence of image post-processing attacks.}
    \label{tab:postprocess}
    \small
    \setlength{\tabcolsep}{2pt}
    \resizebox{\linewidth}{!}{
    \begin{threeparttable}
    \begin{tabular}{llllllllllll}
        \toprule
        Method & P0 & P1 & P2 & P3$^*$  & P4 & P5 & P6 & P7 & P8 & P9$^*$ & P10$^*$ \\
        \midrule
        Stable Sign.     & 0.989 & 0.911 & 0.991 & 0.602 & 0.583 & 0.887 & 0.987 & 0.990 & 0.994 & 0.778 & 0.704 \\
        AquaLoRA         & 0.959 & 0.877 & 0.884 & \underline{0.646} & 0.643 & 0.877 & 0.814 & 0.830 & 0.875 & \underline{0.869} & 0.709 \\
        WMAdapter        & \textbf{0.999} & \underline{0.921} & \textbf{0.994} & 0.639 & \underline{0.651} & \underline{0.973} & 0.988 & 0.992 & 0.997 & 0.651 & \textbf{0.788} \\
        LaWa             & 0.999 & 0.493 & 0.523 & 0.468 & 0.500 & 0.459 & \textbf{0.999} & \textbf{0.999} & \textbf{0.999} & \textbf{0.886} & 0.709 \\
        \rowcolor{cyan!10} EW-LoRA & \underline{0.997} & \textbf{0.922} & \underline{0.993} & \textbf{0.665} & \textbf{0.706} & \textbf{0.983} & \underline{0.989} & \underline{0.992} & \underline{0.997} & 0.849  & \underline{0.781} \\
        \bottomrule
    \end{tabular}
    \begin{tablenotes}
        \footnotesize
        \item The process marked with $*$ means it has been seen in the noise layer.
    \end{tablenotes}
    \end{threeparttable}
    }
\end{table}

\begin{table}[t]
    \centering
    \caption{Watermarking performance in the presence of watermark overwriting attacks.}
    \label{tab:attacks}
    \small
    \setlength{\tabcolsep}{3pt}
    \resizebox{\linewidth}{!}{
    \begin{threeparttable}
    \begin{tabular}{lccccc}
    \toprule
    Attack & Stable Sign. & AquaLoRA & WMAdapter & LaWa & EW-LoRA \\
    \midrule
    StegaStamp & 0.862/0.989 & 0.844/0.959 & 0.941/0.999 & 0.794/0.999 & 0.939/0.997 \\
    SSL-WM.    & 0.988/0.989 & 0.959/0.959 & 0.998/0.999 & 0.999/0.999 & 0.997/0.997 \\
    \bottomrule
    \end{tabular}
    \end{threeparttable}
    }
\end{table}

\subsection{Generalization across Datasets and Models}

In this section, we evaluate the generalization capability of EW-LoRA across different datasets and base models.
Specifically, four datasets, COCO~\cite{linMicrosoftCOCOCommon2014}, CelebA~\cite{liu2015faceattributes}, Flickr~\cite{youngImageDescriptionsVisual2014}, and ArtELingo~\cite{mohamedArtELingoMillionEmotion2022}, are selected to represent diverse visual domains, including everyday scenes, human faces, web imagery, and artistic paintings.
For each dataset, EW-LoRA is trained on one and tested on the others to assess its robustness under cross-domain data distributions. 
In addition, we examine EW-LoRA’s adaptability to different generative architectures by conducting experiments on four base models: Stable Diffusion v1.4 (SD v1.4), Diffusion Transformer (DiT), Stable Diffusion v2.1 (SD v2.1), and Stable Diffusion v3.5-large (SD v3.5-l). All these models adopt a VAE as the image reconstruction backbone, where the VAE decoder maps the latent representations back to the pixel space. Among them, SD v1.4 and DiT share the same VAE decoder consisting of approximately 49.49 MB of parameters, while SD v2.1 and SD v3.5-l utilize slightly larger decoders with 49.63 MB and 49.69 MB of parameters, respectively.

\begin{table}[t]
\centering
\caption{Cross-dataset results. The models are trained on one dataset and test on the others, and their performance is evaluated by BitACC, AT-95 and visual quality metrics PSNR/SSIM/LIPIS/SIFID.}
\setlength{\tabcolsep}{3pt}
\resizebox{\linewidth}{!}{
\begin{tabular}{ll|cc|cccc}
\toprule
Train & Test & BitACC$\uparrow$ & AT-95$\downarrow$ & PSNR$\uparrow$ & SSIM$\uparrow$ & LPIPS$\downarrow$ & SIFID$\downarrow$ \\
\midrule
\multirow{4}{*}{COCO}
  & COCO      & \textbf{0.987} & 1.661 & 32.981 & 0.867 & 0.013 & 0.031 \\
  & CelebA    & 0.957 & 2.253 & \textbf{38.145} & 0.899 & \textbf{0.008} & \textbf{0.014} \\
  & Flickr    & 0.958 & 3.178 & 33.715 & 0.832 & 0.012 & 0.037 \\
  & ArtELingo & 0.968 & \textbf{1.342} & 33.740 & \textbf{0.901} & 0.015 & 0.034 \\
\midrule
\multirow{4}{*}{CelebA}
  & COCO      & 0.982 & 1.485 & 30.084 & 0.792 & 0.028 & 0.083 \\
  & CelebA    & \textbf{0.992} & 2.123 & \textbf{33.791} & 0.818 & \textbf{0.017} & \textbf{0.048} \\
  & Flickr    & 0.966 & 3.154 & 30.669 & 0.729 & 0.026 & 0.095 \\
  & ArtELingo & 0.964 & \textbf{1.175} & 31.010 & \textbf{0.841} & 0.032 & 0.094 \\
\midrule
\multirow{4}{*}{Flickr}
  & COCO      & \textbf{0.992} & \textbf{1.417} & 31.772 & 0.845 & 0.021 & 0.055 \\
  & CelebA    & 0.984 & 1.898 & \textbf{36.177} & \textbf{0.873} & \textbf{0.017} & \textbf{0.035} \\
  & Flickr    & 0.977 & 2.123 & 31.910 & 0.786 & 0.021 & 0.065 \\
  & ArtELingo & 0.978 & 2.124 & 32.136 & 0.872 & 0.025 & 0.064 \\
\midrule
\multirow{4}{*}{ArtELingo}
  & COCO      & \textbf{0.988} & \textbf{1.182} & 32.468 & 0.846 & 0.014 & 0.039 \\
  & CelebA    & 0.977 & 1.987 & \textbf{37.390} & 0.879 & \textbf{0.010} & \textbf{0.022} \\
  & Flickr    & 0.968 & 2.351 & 33.228 & 0.808 & 0.014 & 0.043 \\
  & ArtELingo & 0.972 & 1.456 & 33.513 & \textbf{0.897} & 0.017 & 0.041 \\
\bottomrule
\end{tabular}
}
\label{tab:cross-dataset}
\end{table}

\begin{table}[t]
\centering
\caption{Watermarking performance comparison across different base LDMs}
\setlength{\tabcolsep}{3pt}
\resizebox{\linewidth}{!}{
\begin{tabular}{lccccccc}
\toprule
Model     & Params & BitACC & AT-99 & PSNR & SSIM & LPIPS & SIFID \\
\midrule
SD v1.4   & 0.0353 & 0.997 & 2.931 & 32.543 & 0.869 & 0.035 & 0.093 \\
SD v2.1   & 0.1433 & 0.995 & 2.657 & 32.011 & 0.868 & 0.034 & 0.119 \\
SD v3.5-l & 0.1433 & 0.996 & 2.881 & 32.293 & 0.864 & 0.030 & 0.104 \\
DiT       & 0.0353 & 0.997 & 2.689 & 33.173 & 0.870 & 0.020 & 0.090 \\
\bottomrule
\end{tabular}
}
\label{tab:base_model_comparison}
\end{table}

\emph{Datasets}: As shown in Table~\ref{tab:cross-dataset}, when evaluated across unseen datasets, EW-LoRA exhibits strong generalization, with the embedded watermark remaining largely decodable and the BitACC consistently exceeding 0.95.
Models trained on different datasets show similar robustness when tested on other domains, indicating that the learned watermark representation is not sensitive to data distribution shifts.
In particular, models tested on the CelebA dataset achieve the highest image fidelity, which reflected by higher PSNR ($>$36~dB) and lower LPIPS/SIFID scores, although the watermark accuracy is slightly lower than that of other datasets. Overall, the impact of dataset differences on watermark performance is minor and can be considered negligible, confirming the stable generalization and transferability of EW-LoRA across diverse visual domains.

\begin{figure*}[!tb]
    \centering
    \begin{minipage}[t]{\textwidth}
    \centering
    \includegraphics[width=\textwidth]{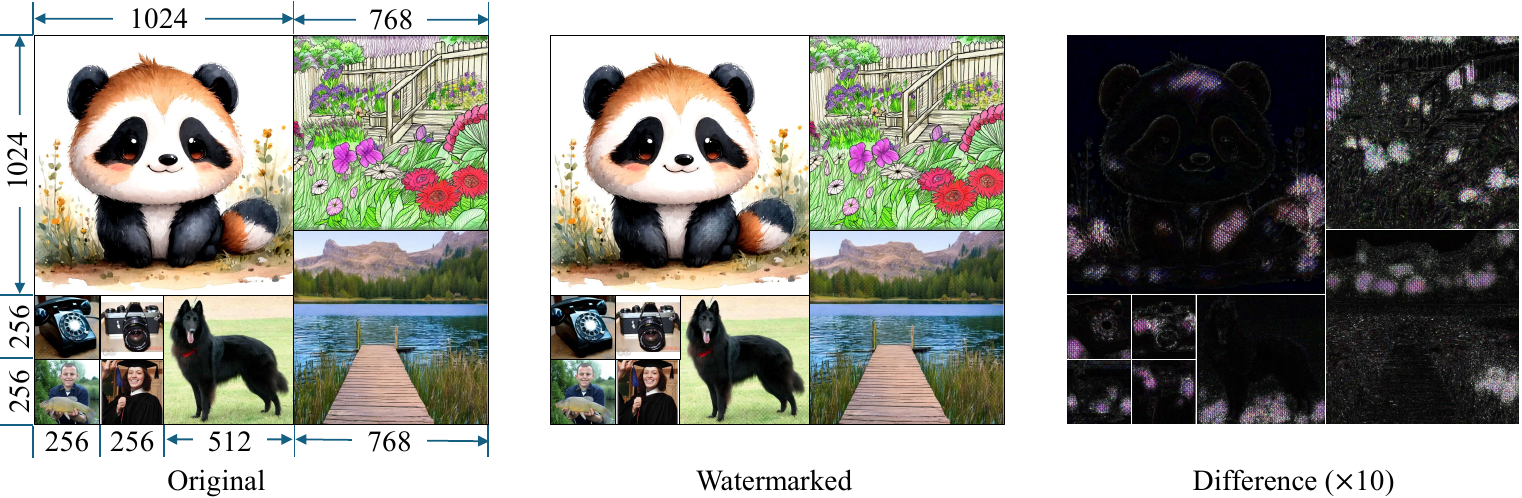}
    \end{minipage}
    \caption{Visualizations for images generated by original LDMs and our EW-LoRA watermarked LDMs, where the images are generated by SD v1.4 (with size of size $256\times 256$), DiT ($512\times 512$), SD v2.1 ($768\times 768$) and SD v3.5-large ($1024\times 1024$).}
    \label{fig:genimgs_diff_models}
\end{figure*}

\emph{Base Models}: As shown in Table~\ref{tab:base_model_comparison}, we apply EW-LoRA to different base models for watermarking.
The results show that even when the underlying generative architectures differ substantially, ranging from the U-Net backbone used in Stable Diffusion to the transformer-based structure adopted by DiT, EW-LoRA consistently achieves high watermark accuracy (BitACC~$\geq$~0.995) and comparable image quality across all models.
This demonstrates that EW-LoRA is architecture-agnostic and can be seamlessly integrated into diverse generative frameworks without compromising watermark robustness or visual imperceptibility.
Figure~\ref{fig:genimgs_diff_models} further illustrates representative samples generated by SD v1.4 ($256\times256$), SD v2.1 ($768\times768$), DiT ($512\times512$), and SD v3.5-l ($1024\times1024$). From the difference images, we observe that the watermark is primarily distributed within texture-rich and detail-dense regions of the generated images.
These regions provide abundant local variations that effectively conceal the watermark signal, ensuring that the embedded watermark remains visually imperceptible to the human eye (invisible watermarking).

\subsection{Analysis of DLWS}

\begin{figure}[!tb]
    \centering
    \begin{minipage}[t]{\columnwidth}
    \centering
    \includegraphics[width=\columnwidth]{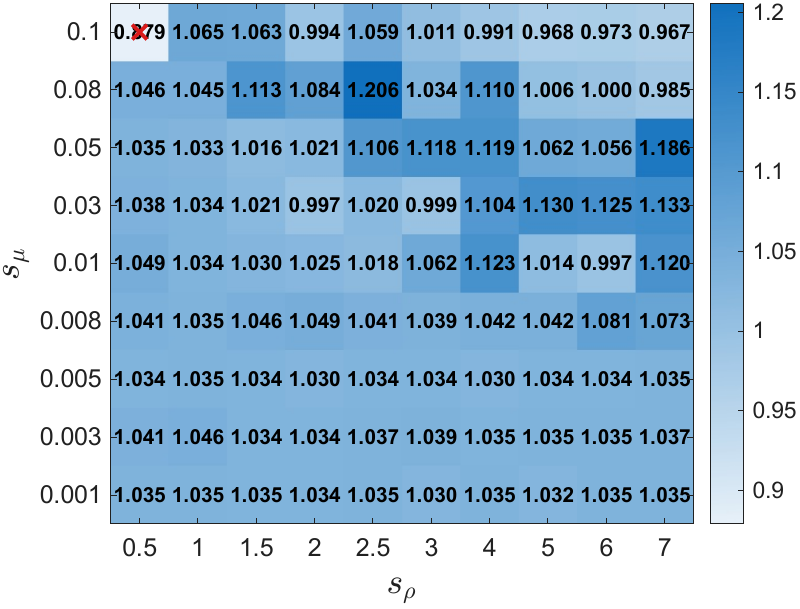}
    \end{minipage}
    \caption{A heatmap showing the scores for evaluating the optimal watermarking performance, based on the combinations of the step factors $s_\rho, s_\mu$.} 
    \label{fig:hyper_params_analysis}
\end{figure}

In the proposed Dynamic Loss Weight Scheduler (DLWS), as detailed in Algorithm~\ref{algo:dlwt}, the targets for PSNR and BitACC are adjusted dynamically throughout the training process. Initially, these thresholds are set conservatively (typically below 40 dB for PSNR and 0.95 for BitACC) to prevent overfitting one objective while hindering the convergence of the other.
Once the validation performance reaches these preset targets, the thresholds are gradually increased by step factors $s_\rho$ and $s_\mu$ to further challenge the model. If these targets are consistently met over several iterations, the model is considered converged.
Figure~\ref{fig:hyper_params_analysis} illustrates the impact of different combinations of $(s_\rho, s_\mu)$, with the values assessed based on the score of $\text{norm(PSNR)}+\text{BitACC}$. The best combination is selected by choosing the pair that yields the highest value. From the figure, we can select the best parameter values ($s_\rho,s_\mu$) = ($2.5, 0.08$).

\section{Conclusion}

In summary, the proposed EW-LoRA framework demonstrates that watermarking for latent diffusion models can be achieved in a highly efficient and unobtrusive manner. By embedding watermarks into the VAE decoder through low-rank adaptation, EW-LoRA preserves generation fidelity while significantly reducing the number of trainable parameters. The incorporated dynamic loss weight scheduler (DLWS) enables faster and more stable convergence by adaptively balancing the objectives of image generation and watermark embedding. As a general and lightweight optimization mechanism, DLWS also offers potential for broader application across various watermarking paradigms.
Future work will explore finer layer selection strategies within diffusion components, investigate adaptive rank allocation for LoRA modules, and extend the framework to other generative architectures such as autoregressive-based diffusion models for a unified watermarking solution.

\bibliographystyle{elsarticle-harv}
\bibliography{sd-lora-wm}

\end{document}